\documentclass{article} 
\usepackage{longcat_style,times}

\usepackage{amsmath,amsfonts,bm}

\def\eqref#1{equation~\ref{#1}}

\def\1{\bm{1}}

\DeclareMathAlphabet{\mathsfit}{\encodingdefault}{\sfdefault}{m}{sl}
\SetMathAlphabet{\mathsfit}{bold}{\encodingdefault}{\sfdefault}{bx}{n}

\usepackage{hyperref}
\usepackage{url}

\usepackage{graphicx}
\usepackage{balance}  
\usepackage{amsmath}
\usepackage{algpseudocode}
\usepackage{latexsym}
\usepackage{multirow}
\usepackage{multicol}
\usepackage{color}
\usepackage{dashrule}
\usepackage{extarrows}
\usepackage{dsfont}
\usepackage{centernot}
\usepackage{hyperref}
\usepackage{natbib}
\usepackage{fancybox}
\usepackage[utf8]{inputenc} 
\usepackage[T1]{fontenc}    
\usepackage{hyperref}       
\usepackage{url}            
\usepackage{booktabs}       
\usepackage{amsfonts}       
\usepackage{amsmath}
\usepackage{nicefrac}       
\usepackage{microtype}      
\usepackage{xcolor} 
\usepackage{amsmath}
\usepackage{lipsum}
\usepackage{colortbl}
\usepackage{subcaption}
\usepackage{tabularx}
\usepackage{makecell}
\usepackage{wrapfig}
\usepackage{bm}
\usepackage{multirow}
\usepackage{xcolor} 
\usepackage{bbm}
\usepackage{mdframed}
\usepackage{enumitem}
\usepackage{xspace}

\usepackage{lineno}

\definecolor{darkblue}{rgb}{0, 0, 0.5}
\hypersetup{colorlinks=true, citecolor=darkblue, linkcolor=darkblue, urlcolor=darkblue}

\usepackage{listings}
\usepackage{xcolor}
\usepackage[ruled,vlined]{algorithm2e}
\usepackage{caption}
\usepackage{float} 
\usepackage{arydshln}
\usepackage{wrapfig} 
\usepackage{tcolorbox}
\tcbuselibrary{theorems}
\tcbuselibrary{most}
\usepackage[dvipsnames]{xcolor}

\definecolor{customTeal}{RGB}{0, 128, 128} 
\definecolor{emphasisColor}{RGB}{255, 0, 0} 
\definecolor{oursBlue}{RGB}{51,202,246}
\newcommand{\name}{\textsc{R-Horizon}\xspace}
\definecolor{blue1}{HTML}{508AB2}
\definecolor{green2}{HTML}{BFF6BA}

\newtcbtheorem[]{prompt}{Prompt}%
{colback=SeaGreen!10!CornflowerBlue!10,
 colframe=RoyalPurple!55!Aquamarine!100!,
 fonttitle=\bfseries,
 left=.02in, right=.02in, bottom=.02in, top=.02in,
 before upper={\linespread{1.5}\selectfont}}{prompt}

\renewcommand{\headeright}{\raisebox{-0.2\height}{\includegraphics[height=1.8em]{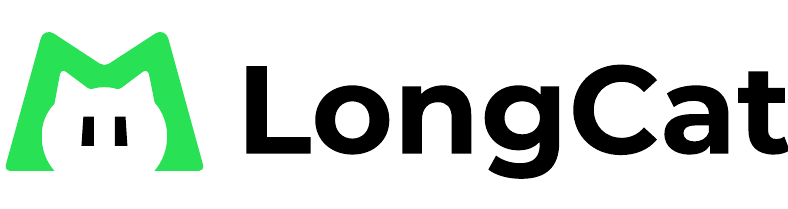}}}
\renewcommand{\shorttitle}{How Far Can Your Large Reasoning Model Really Go in Breadth and Depth?}

\definecolor{darkblue}{rgb}{0, 0, 0.5}
\hypersetup{colorlinks=true, citecolor=darkblue, linkcolor=darkblue, urlcolor=darkblue}

\makeatletter
\renewcommand{\@maketitle}{%
  \vbox{%
    \hsize\textwidth
    \linewidth\hsize
    \vskip -0.5in
    \noindent
    \begin{minipage}{0.2\textwidth}
      \includegraphics[width=\linewidth]{figures/longcat-logo-full.pdf}
    \end{minipage}%
    \\
    \rule{\linewidth}{1pt}
    \hspace{0.05\textwidth}%
    \begin{minipage}{0.8\textwidth}
    \end{minipage}

    \centering
    {\LARGE \bfseries\@title\par}
    \vskip 0.1in  
    \def\And{%
      \end{tabular}\hfil\linebreak[0]\hfil%
      \begin{tabular}[t]{c}\bf\rule{\z@}{24\p@}\ignorespaces%
    }
    \def\AND{%
      \end{tabular}\hfil\linebreak[4]\hfil%
      \begin{tabular}[t]{c}\bf\rule{\z@}{24\p@}\ignorespaces%
    }
    \begin{tabular}[t]{c}\bf\rule{\z@}{24\p@}\@author\end{tabular}%
  \vskip 0.05in 
  }
}
\makeatother

\title{\name: How Far Can Your Large Reasoning Model Really Go in Breadth and Depth? \\}

\makeatletter
\def\@fnsymbol#1{\ensuremath{\ifcase#1\or \dagger\or \ddagger\or
   \mathsection\or \mathparagraph\or \|\or **\or \dagger\dagger
   \or \ddagger\ddagger \else\@ctrerr\fi}}
\makeatother

\author{
\begin{tabular}{c}
\textbf{Yi Lu}$^{1,2}$
\quad
\textbf{Jianing Wang}$^{2}$\thanks{Corresponding authors. Correspondence to: 
lygwjn@gmail.com, wangwei432@meituan.com}
\quad
\textbf{Linsen Guo}$^{2}$
\quad
\textbf{Wei He}$^{1,2}$
\quad
\textbf{Hongyin Tang}$^{2}$ \\[1ex]
\textbf{Tao Gui}$^{1}$
\quad
\textbf{Xuanjing Huang}$^{1}$
\quad
\textbf{Xuezhi Cao}$^{2}$
\quad
\textbf{Wei Wang}$^{2}$\footnotemark[1]
\quad
\textbf{Xunliang Cai}$^{2}$ \\[1ex]
\normalfont $^1$Fudan University 
\quad
$^2$Meituan LongCat Team\\[1ex]
\normalfont \texttt{\url{https://github.com/meituan-longcat/R-HORIZON}}
\end{tabular}
}

\begin{document}

\maketitle

\begin{abstract}
Recent trends in test-time scaling for reasoning models (e.g., OpenAI o1, DeepSeek-R1) have led to remarkable improvements through long Chain-of-Thought (CoT). However, existing benchmarks mainly focus on immediate, single-horizon tasks, failing to adequately evaluate models' ability to understand and respond to complex, long-horizon scenarios. To address this incomplete evaluation of Large Reasoning Models (LRMs), we propose \name, a method designed to stimulate long-horizon reasoning behaviors in LRMs through query composition. Based on \name, we construct a long-horizon reasoning Benchmark, comprising complex multi-step reasoning tasks with interdependent problems that span long reasoning horizons. Through comprehensive evaluation of LRMs using the \name Benchmark, we find that even the most advanced LRMs suffer significant performance degradation. Our analysis reveals that LRMs exhibit limited effective reasoning length and struggle to allocate thinking budget across multiple problems appropriately. Recognizing these limitations, we use \name to construct long-horizon reasoning data for reinforcement learning with verified rewards (RLVR). Compared to training with single-horizon data, RLVR with \name not only substantially improves performance on the multi-horizon reasoning tasks, but also promotes accuracy on standard reasoning tasks (+7.5 on AIME2024).
These results position \name as a scalable, controllable, and low-cost paradigm for enhancing and evaluating the long-horizon reasoning capabilities of LRMs.

\end{abstract}

\begin{figure*}[ht]
  \centering
  \includegraphics[width=\textwidth, trim=0 0 0 0, clip] {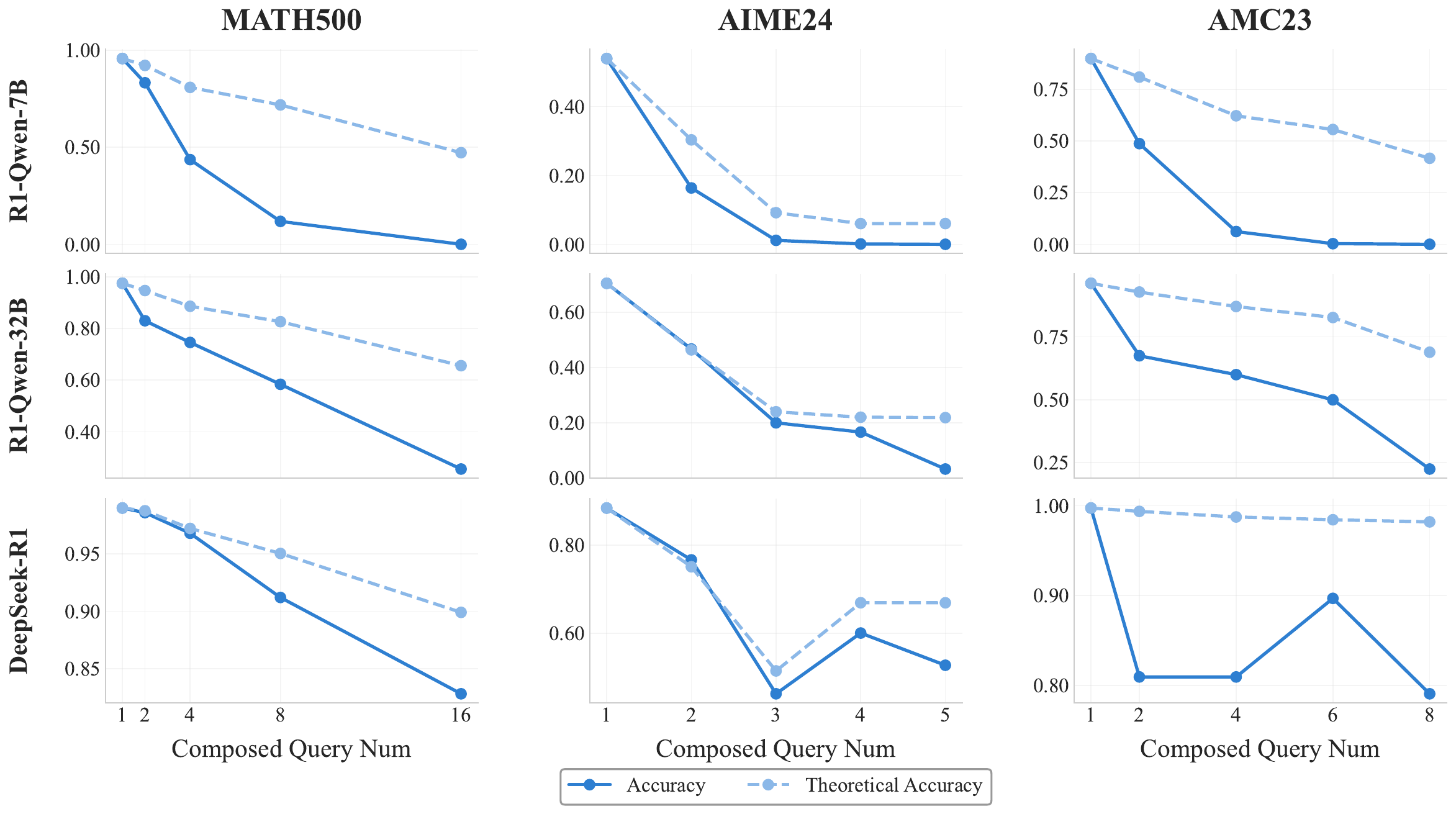}
  \caption{Actual versus theoretical accuracy of R1-series models on \name datasets. 
  }
  \label{fig:mainfig}
  \vspace{-.5em}
\end{figure*}

\section{Introduction}

Recent advances in reasoning-focused language models, exemplified by OpenAI's o1~\citep{openai2024openaio1card} and DeepSeek-R1~\citep{guo2025deepseek}, have established test-time scaling as a fundamental component for enhancing reasoning abilities in large reasoning models (LRMs). Specifically, test-time scaling enables long Chain-of-Thought (CoT) and induces sophisticated reasoning behaviors, leading to remarkable improvements on challenging reasoning tasks like mathematical reasoning~\citep{he2025deepmath103k,yu2025dapoopensourcellmreinforcement,yue2025vapoefficientreliablereinforcement,zeng2025simplerlzooinvestigatingtamingzero}, code generation~\citep{deepcoder2025, zeng2025simplerlzooinvestigatingtamingzero} and agentic tasks~\citep{kimiteam2025kimik2openagentic,5team2025glm45agenticreasoningcoding}. 

By continuously expending computational resources throughout the reasoning process, models with longer reasoning trajectories achieve superior performance on various reasoning benchmarks~\citep{muennighoff_s1_2025}, yet this simultaneously exposes critical limitations in current training and evaluation paradigms. 
Existing training and evaluation datasets~\citep{gsm8k,math500,jain2024livecodebench} primarily confine themselves to the reasoning of isolated problems, focusing on immediate single-horizon tasks where questions and answers remain independent of each other. 
However, real-world scenarios often require an AI agent to reason, plan, and act over an extended series of steps, sometimes thousands or even millions, where inference must span across multiple sequential and potentially interdependent problems~\citep{yao2024taubench,tao2025webshaper}. Single-horizon evaluation paradigms cannot effectively assess the ability of a model to understand and respond to complex, multi-horizon tasks or scenarios that require a sequence of logical steps over a longer period of time. 
Moreover, conventional reinforcement learning (RL) typically focuses on single, isolated problems, preventing models from developing long-horizon reasoning capabilities to tackle multiple problems through the RL process. 
The incomplete picture of training and evaluation paradigms raises a fundamental question: \textit{How far can large reasoning models really go in breadth and depth?}




In this study, we propose \name, a simple yet effective method to stimulate long-horizon reasoning behaviors in LRMs through query composition. 
This method aims to construct dependencies and concatenate existing single-horizon tasks, transforming isolated problems into complex multi-horizon reasoning scenarios. 
For instance, in mathematical tasks, we first extract key information from all problems, then establish dependencies by linking one problem's answer to another problem's critical information, requiring models to solve multiple problems sequentially to obtain all correct answers. 
To address the limitations of current training and evaluation paradigms,
we leverage this method to establish an evaluation benchmark and training data to evaluate and enhance the long-horizon reasoning capabilities of LRMs.

We first establish \name benchmark, which comprises 6 representative datasets across mathematics, code generation, and agent applications  (e.g. MATH500~\citep{math500}, LiveCodeBench~\citep{jain2024livecodebench}, WebShaper~\citep{tao2025webshaper}).
Through evaluating 25 mainstream LRMs, we find that even the most advanced LRMs suffer significant performance degradation on \name benchmark. The performance in multi-horizon reasoning scenarios falls substantially below the theoretical performance (Figure~\ref{fig:mainfig}). 
Further analysis reveals critical limitations of current LRMs that contribute to the performance gap: 
(1) LRMs possess a limited effective reasoning length, with performance declining sharply once the thinking budget exceeds this threshold.
(2) LRMs exhibit constrained reflection scope—LRMs often reflect within the current problem, failing to identify errors from previous questions. 
(3) The overthinking phenomenon~\citep{chen2025think23overthinkingo1like} in LRMs prevents the appropriate allocation of thinking budget when facing multiple reasoning problems.

To address the absence of long-horizon problems in current RL training data, we leverage \name to reconstruct training datasets and design different reward functions, then conduct reinforcement learning with verified rewards (RLVR) with varying composed problems and reward schemes to investigate the impact of long-horizon reasoning data on the training process.
By employing mainstream RLVR algorithms GRPO~\citep{GRPOshao2024deepseekmath} with \name{}, we observe that traditional RLVR provides limited improvements on multi-step reasoning tasks. In contrast, training with \name{} data is a highly efficient training approach that not only enhances single-problem performance more effectively but also rapidly improves performance on multiple problems. 
Our analysis demonstrates that training with \name also improves response length efficiency and thinking budget allocation. 
In summary, \name{} mitigates the current limitations of long-horizon reasoning in training and evaluation paradigms—offering a scalable, controllable and low-cost path to improve and evaluate the long-horizon abilities of LRMs.

\section{Related Work}
\vspace{-.5em}
\subsection{Test Time Scaling in Large Reasoning Models}
\vspace{-.5em}
The success of OpenAI's o1 introduced a new scaling paradigm, test-time compute scaling, which improves performance through increasing inference computation~\citep{openai2024openaio1card}. 
However, recent studies reveal that LRMs may generate verbose reasoning trajectories with marginal accuracy gains. \citet{chen2025think23overthinkingo1like} reveals the ``overthinking'' phenomenon, showing that LRMs generate significantly more tokens than conventional LLMs on simple arithmetic tasks, with minimal increase in accuracy.
To address this, \citet{aggarwal2025l1controllinglongreasoning} proposed length-controlled policy optimization, providing precise control over the length of the reasoning trajectories during generation. \citet{yang2025thinkingoptimalscalingtesttimecompute} developed a thinking-optimal scaling strategy, allowing models to flexibly adjust their reasoning depth according to the available test-time compute budget. 
Recent studies have also focused on fine-tuning models to think efficiently according to task complexity~\citep{hao2024traininglargelanguagemodels,liu2025learnreasonefficientlyadaptive,fang2025thinklessllmlearnsthink,arora2025traininglanguagemodelsreason,zhang2025adaptthinkreasoningmodelslearn}.
In agentic tasks, overthinking also reduces performance while increasing inference costs~\citep{cuadron2025dangeroverthinkingexaminingreasoningaction}.
Although previous studies indicate that overthinking leads to computational inefficiency with limited performance gains, our findings reveal that prolonged reasoning substantially degrades performance on compound multi-step reasoning tasks.

\subsection{Effective Reasoning Length of Large Reasoning Models}
\vspace{-.5em}
Recent studies explore the effective reasoning length of LRMs in mathematical benchmarks (i.e., GSM8k~\citep{gsm8k}, MATH500~\citep{math500} and AIME~\citep{AIME2024,AIME2025}).
\citet{su2025underthinkingoverthinkingempiricalstudy,yang2025thinkingoptimalscalingtesttimecompute,wu2025lessunderstandingchainofthoughtlength} investigates the relationship between reasoning length and accuracy. \citet{su2025underthinkingoverthinkingempiricalstudy} finds that models fail to adaptively calibrate their response length according to the problem difficulty.
\citet{wu2025lessunderstandingchainofthoughtlength, ghosal2025doesthinkinghelpunderstanding,chen2024unlockingcapabilitiesthoughtreasoning} demonstrate the existence of an optimal CoT length beyond which performance degrades.
By directly concatenating multiple independent questions, REST~\citep{pan2025reststresstestinglarge} reveals that LRMs fail to keep their performance under multi-context stress.
However, these tasks either focus on a single problem or concatenate independent problems without meaningful logical dependencies. In contrast, we design multi-dependent synthetic tasks to expose failure modes amplified by extended reasoning, consistent with findings that reasoning chains exceeding optimal length reduce accuracy.

\vspace{-.5em}
\section{\name}
\begin{figure*}[ht]
  \centering
  \includegraphics[width=\textwidth]{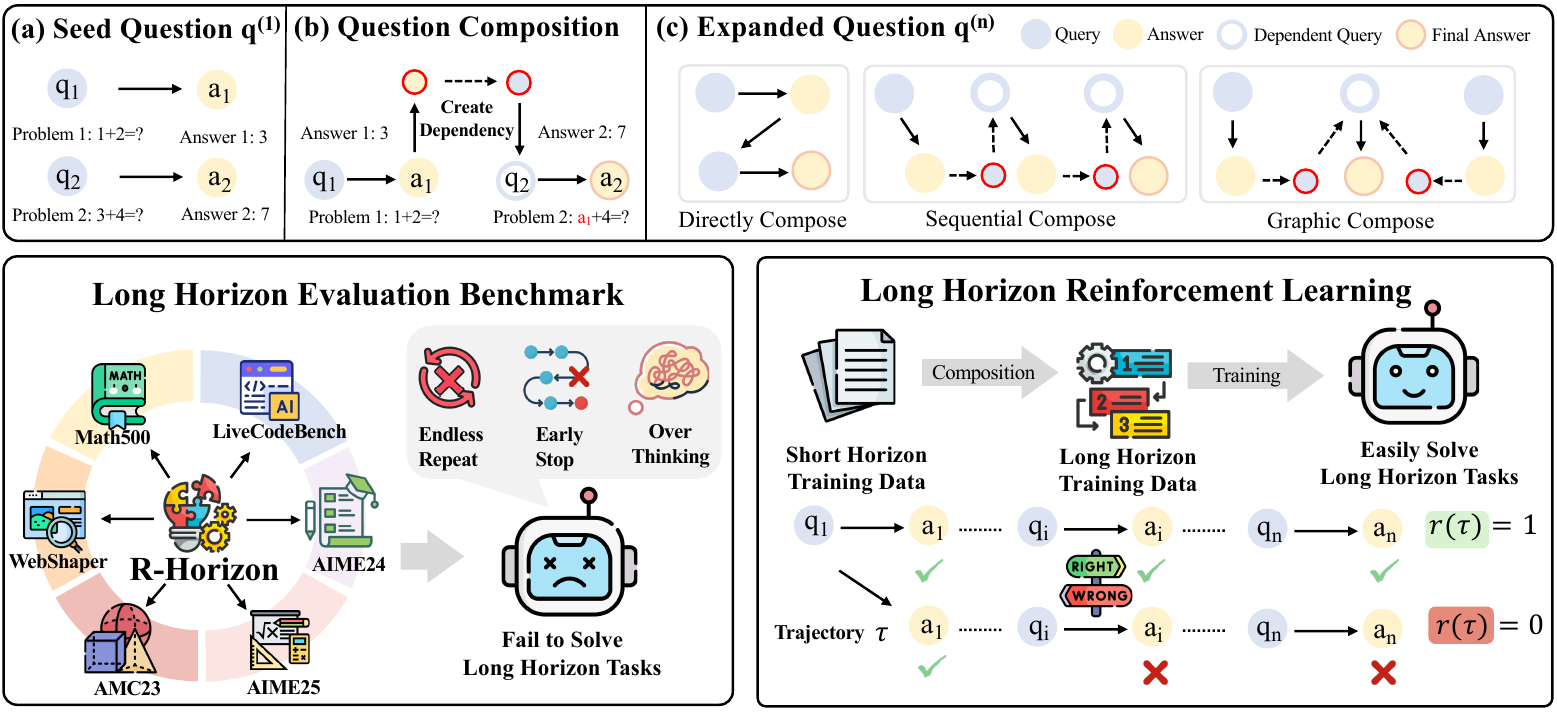}
  \caption{The \name data composition pipeline is illustrated in (a)-(c). We leverage \name to construct a comprehensive long-horizon reasoning evaluation benchmark spanning 6 tasks and generate multi-horizon training data for long-horizon reinforcement learning.
  }
  \label{fig:method}
  \vspace{-1em}
\end{figure*}

\vspace{-.5em}
We propose \name, a method designed to stimulate long-horizon reasoning behaviors in LRMs via query composition. As illustrated in Figure~\ref{fig:method}, \name supports the concatenation of three types of expanded questions and can be employed in both the training and evaluation stages to enhance and evaluate the long-horizon capabilities of LRMs.

\subsection{\name Datasets Construction}
\vspace{-.5em}


For mathematical tasks, we adopt the \textit{sequentially composed} concatenation to construct a dataset of multi-step mathematical problems with explicit dependencies that enforce sequential solving. The construction pipeline consists of two stages: seed problem filtering and expanded problem composition. For code and agentic tasks, we provide the construction process in Appendix~\ref{appendix:dataset_construction}.
\vspace{-.5em}
\paragraph{Seed Problem Filtering}
\label{method:problem_filtering}
Given an initial dataset $\mathcal{D} = \{(q_i, a_i)\}_{i=1}^N$ 
where $(q_i, a_i)$ is a pair of a question and an answer.
We apply the following filtering criteria to obtain a seed set $\mathcal{D}_{\text{seed}}$:
\begin{equation}
\mathcal{D}_{\text{seed}} = \left\{(q, a) \in \mathcal{D} \mid \left|I(q)\right| > 0 \land a \in \mathbb{Z}\right\},
\end{equation}
where $I(\cdot) = \texttt{extract\_int}(\cdot)$ denotes extracting all integers appearing in the input text.

For each $(q, a) \in \mathcal{D}_{\text{seed}}$, we identify key variables from the extracted integers. 
We then employ a model $M$ to verify each interger $m \in I(q)$ whether is a key variable:
\begin{equation}
K(q) = \left\{m \in I(q) \mid M(q, m) = 1\right\},
\end{equation}
where $M(q, m) = 1$ indicates that removing $m$ from $q$ renders the problem unsolvable. 
Each filtered seed problem is then represented as a triple $(q, a, K(q))$.

\paragraph{Expanded Problem Composition}
\label{method:query_composition}
Given seed problems with annotated key variables, we construct dependency chains using Algorithm~\ref{alg:compose}. The algorithm ensures that each modified problem $q_{i+1}'$ contains a placeholder variable $v_{i+1}$ that must be resolved through the dependency function $f_i(a_i) = m_{i+1}$, requiring the solution $a_i$ from the previous problem. The augmentation step prepends the dependency specification to the problem statement, making the sequential constraint explicit. The final dataset $\mathcal{D}_{\text{composed}}$ consists of problem sequences that enforce strict sequential solving. 

\vspace{-.5em}
\begin{algorithm}[ht]
\SetAlgoLined
\KwIn{Seed problems $\{(q_1, a_1, K_1), \ldots, (q_n, a_n, K_n)\}$}
\KwOut{Composed problem $\mathcal{Q}$}
Initialize $\mathcal{Q} \leftarrow [q_1]$\;
\For{$i = 1$ to $n-1$}{
    Select key variable $m_{i+1} \in K_{i+1}$\ and Create placeholder variable $v_{i+1}$\;
    
    Define dependency function $f_i(x) \leftarrow x + (m_{i+1} - a_i)$\;
    Substitute $m_{i+1}$ with $v_{i+1}$ in $q_{i+1}$ to obtain $q_{i+1}'$\;
    Augment $q_{i+1}'$ with dependency constraint $v_{i+1} = f_i(a_i)$\;
    Append $q_{i+1}'$ to $\mathcal{Q}$\;
}
\Return $\mathcal{Q} = (q_1, q_2', \ldots, q_n')$\;
\caption{Dependency Chain Construction}
\label{alg:compose}
\end{algorithm}
\vspace{-.5em}

\subsection{\name Benchmark}
\vspace{-.5em}
We use \name{} to reconstruct existing evaluation datasets, combining different datasets through problem filtering and composition approaches, and design evaluation metrics for composed problems.

\vspace{-.5em}
\paragraph{Evaluation Metrics}~\name{} evaluates model performance by extracting all answers from the model's response. Given a composed problem sequence $\mathcal{Q} = (q_1, q_2', \ldots, q_n') \in \mathcal{D}_{\text{composed}}$, we extract the corresponding answer sequence $\hat{\mathcal{A}} = (\hat{a}_1, \hat{a}_2, \ldots, \hat{a}_n)$ from the model's response $\mathcal{R}$. 
We use all-or-nothing scoring: correct only if all sub-problems are solved:
\begin{equation}
\text{Acc}(\mathcal{Q}) = \begin{cases}
1 & \text{if } \hat{a}_i = a_i \text{ for all } i \in \{1, \ldots, n\}, \\
0 & \text{otherwise.}
\end{cases}
\end{equation}
We also propose a metric to estimate its theoretical accuracy. For each $(q, a) \in \mathcal{D}_{\text{seed}}$, we use the pass rate of these atomic problems to estimate the expected accuracy of composed problems through:
\begin{equation}\label{eq:expected-acc}
\text{Acc}_{\text{expected}}(\mathcal{Q}) = \prod_{i=1}^{n} p_i,
\end{equation}
where $p_i$ is the pass rate of atomic problem $q_i$. We use model-based extraction to handle diverse response formats (details in Appendix~\ref{appendix:evaluation_metrics}).

\subsection{Reinforcement Learning with \name{}}
\vspace{-.5em}
To investigate the formation of long-horizon reasoning capabilities and understand how multi-step dependent queries influence the reinforcement learning process, we employ \name{} datasets as training data for reinforcement learning from verifiable rewards (RLVR). We follow Skywork-OR1~\citep{skywork-or1-2025} RLVR pipelines while utilizing our constructed training data.
\vspace{-.5em}
\paragraph{Group Relative Policy Optimization (GRPO)}
We adopt GRPO~\citep{GRPOshao2024deepseekmath} as our optimization algorithm, which eliminates the value function requirement of PPO~\citep{PPO} by computing advantages in a group-relative manner. 
For each question $q$, the behavior policy $\pi_{\theta_\text{old}}$ samples a group of $G$ response candidates $\{o_1, \ldots, o_G\}$. We use GRPO with token-level policy gradient loss, which optimizes the policy model by maximizing the following objective:
\begin{equation}
\resizebox{\textwidth}{!}{$
\begin{aligned}
    \mathcal{J}&_{\text{GRPO}}(\theta) = \mathbb{E}_{q, \{o_i\}_{i=1}^G} \frac{1}{\sum_{i=1}^G |o_i|}\sum_{i=1}^G \sum_{t=1}^{|o_i|} \left\{ \min \left(r_{i,t} \hat{A}_{i,t},\text{clip} \left( r_{i,t}, 1 - \epsilon, 1 + \epsilon \right)\hat{A}_{i,t} \right) - \beta \mathbb{D}_{\text{KL}}\left[\pi_{\theta} || \pi_{\text{ref}}\right]\right\} ,
\end{aligned}
$}
\label{eq:GRPO-obj}
\end{equation}
where $r_{i,t}=\frac{\pi_\theta(o_{i,t} | q, o_{i,<t})}{\pi_{\theta_{\text{old}}}(o_{i,t} | q, o_{i,<t})}$, $\epsilon$ and $\beta$ are hyperparameters, $\hat{A}_{i,t}$  is the advantage calculated based on the relative rewards of the outputs inside each group only, and $\mathbb{D}_{\text{KL}}$ denotes the KL divergence between the learned policy and a reference policy $\pi_{\text{ref}}$.

\vspace{-.5em}
\paragraph{Reward Design}
We design two reward schemes for multi-horizon training data:

\begin{equation}
R_{\text{last}} = \begin{cases}
1 & \text{if } \hat{a}_n = a_n, \\
0 & \text{otherwise},
\end{cases}
\quad \text{and} \quad
R_{\text{all}} = \begin{cases}
1 & \text{if } \hat{a}_i = a_i \text{ for all } i \in \{1, \ldots, n\}, \\
0 & \text{otherwise}.
\end{cases}
\end{equation}

Last-only reward $R_{\text{last}}$ provides feedback on the final answer only, while all-correct reward $R_{\text{all}}$ requires all intermediate steps to be correct. 
This distinction allows us to study how different reward function influence the development of long-horizon reasoning capabilities.

\section{Experiment}
\vspace{-.5em}
\subsection{Evaluation Setup}
\vspace{-.5em}
\paragraph{Datasets}

For mathematical tasks, we construct MATH500~\citep{math500}, AIME24~\citep{AIME2024}, and AIME25~\citep{AIME2025} with multiple dependent queries, using $n \in \{1, 2, 4, 8, 16\}$ for MATH500 and $n \in \{1, 2, 3, 4, 5\}$ for the more challenging AIME datasets. For code tasks, we reconstruct LiveCodeBench (v5)~\footnote{The time is ranged from August 2024 to May 2025.}~\citep{jain2024livecodebench} with $n \in \{1, 2, 3, 4, 5\}$. For agentic tasks, we use WebShaper~\citep{tao2025webshaper} with multi-round tool calls for web search ($n \in \{1, 2, 3, 4, 5\}$). See Appendix~\ref{appendix:model_datasets_details} for details.

\paragraph{Models}

\vspace{-.5em}
We select 25 advanced LRMs to perform evaluation on our built \name benchmark,
including the R1-distill series models~\citep{guo2025deepseek}, Qwen series~\citep{yang2025qwen3technicalreport} models, and Nemotron~\citep{bercovich2025llamanemotronefficientreasoningmodels} series models. 
Model details are in Appendix~\ref{appendix:model_datasets_details}. We set the generation length to 64k tokens to avoid truncation. More inference settings are in Appendix~\ref{appendix:evaluation_inference_setting}.

\subsection{Evaluation Result}
\vspace{-.5em}

\begin{figure*}[ht]
  \centering
  \includegraphics[width=\textwidth]{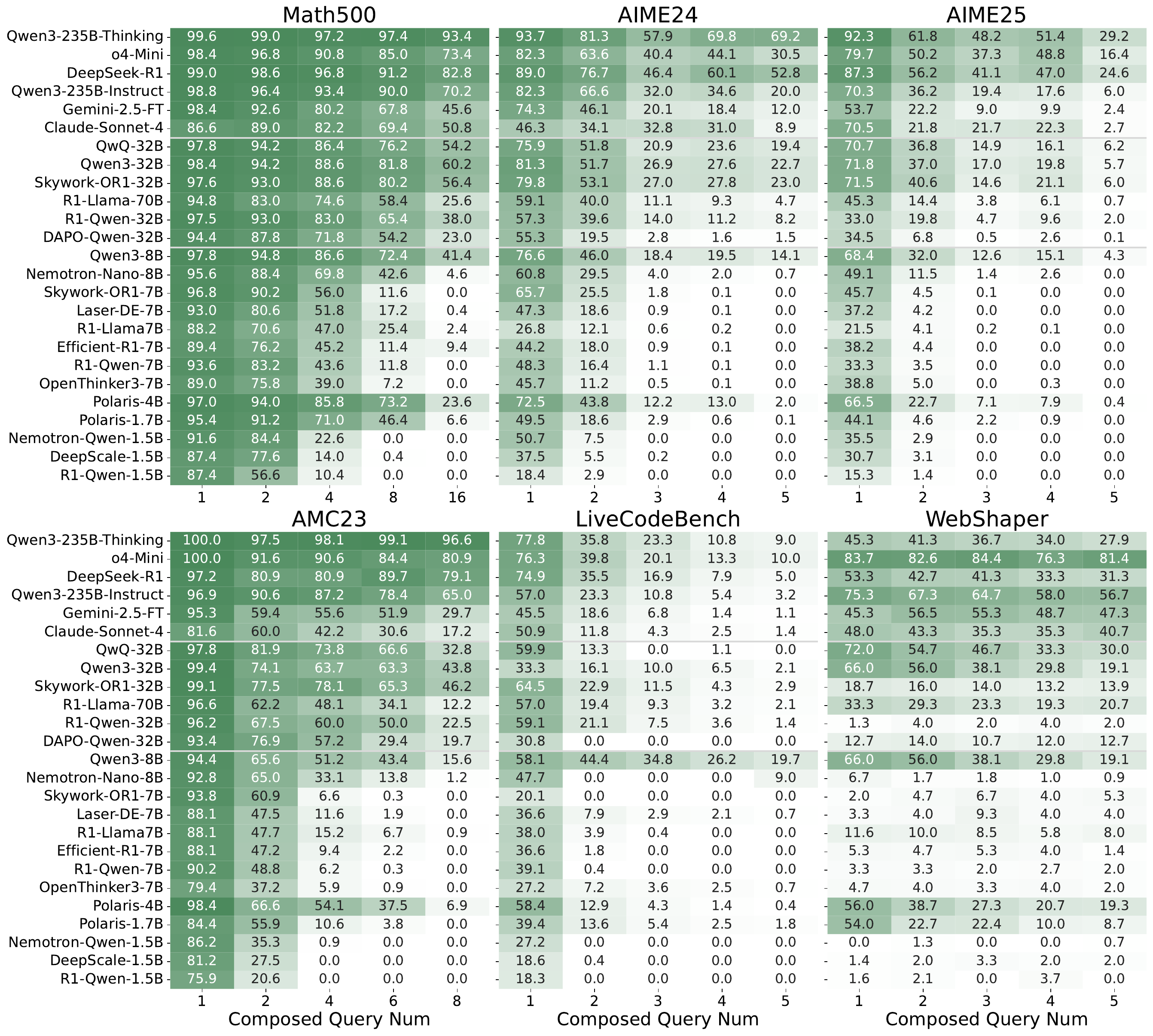}
  \caption{Evaluation results of \name Benchmark.}
  \label{fig:result}
  \vspace{-1em}
\end{figure*}

\paragraph{Performance Degradation as the Reasoning Horizon Increases}
As shown in Figure~\ref{fig:result}, we observe that models across different categories experience performance degradation as the reasoning horizon increases. 
Even the most powerful models, including DeepSeek-R1, Qwen3-235B-A22B-Thinking, and o4-mini, suffer from severe performance degradation as the reasoning horizon increases. For instance, on AIME25, DeepSeek-R1 drops from 87.3\% ($n=1$) to 24.6\% ($n=5$). 
Additionally, we find that larger models exhibit less degradation when confronting composed problems, while smaller models experience more severe performance degradation. For example, R1-Qwen-7B drops from 93.6\% ($n=1$) to 0\% ($n=16$), which is 34.1\% more than the 32B model. 

\paragraph{Same Degradation Trends Across Different Model and Task Categories}
We observe consistent degradation trends across tasks of varying difficulty and types. Models exhibit greater performance drops when facing more challenging tasks. For instance, Qwen3-235B-Thinking drops from 93.7\% ($n=1$) to 69.2\% ($n=5$) on AIME24, but experiences a steeper decline from 92.3\% ($n=1$) to 29.2\% ($n=5$) on AIME25. For code tasks, we find that the degradation trend is more severe compared to mathematical tasks, with smaller models (7B) struggling to complete multiple code problems. For web search tasks, we observe that many trained reasoning models have lost their ability to call tools, resulting in poor performance.

\subsection{Reinforcement Learning with \name Datasets}
\label{sec:training_horizon}
\vspace{-.5em}
Despite reinforcement learning bringing long CoT thinking capabilities to models, current mainstream LRMs still cannot achieve good performance on \name{} Benchmark. 
We follow Skywork-OR1~\citep{skywork-or1-2025} to observe the changes in long-horizon reasoning capabilities of long CoT models before and after standard RL in Appendix~\ref{appendix:skywork_baseline}.
We find that training with only single-problem data leads to slow improvement in models' ability to handle composed problems. To investigate the impact of \name{} data on RL training, we construct composed training data through \name based on the original math training datasets. 

\paragraph{Training Setup}
We construct a data pool $\mathcal{D}_{\text{filtered}}$ from Skywork-OR1-RL training data using Problem Filtering (Section~\ref{method:problem_filtering}). To manage difficulty, we combine problems by pass rates, keeping $\text{Acc}_{\text{expected}} > 0.25$. We train on R1-Qwen-7B and set maximum response length to 40k to prevent truncation and use the last-only reward $R_{\text{last}}$ as default, which provides feedback on the final answer only. Details are in Appendix~\ref{appendix:training_setup}.

\paragraph{Training with \name Datasets}

\begin{figure*}[ht]
  \centering
  \includegraphics[width=\textwidth]{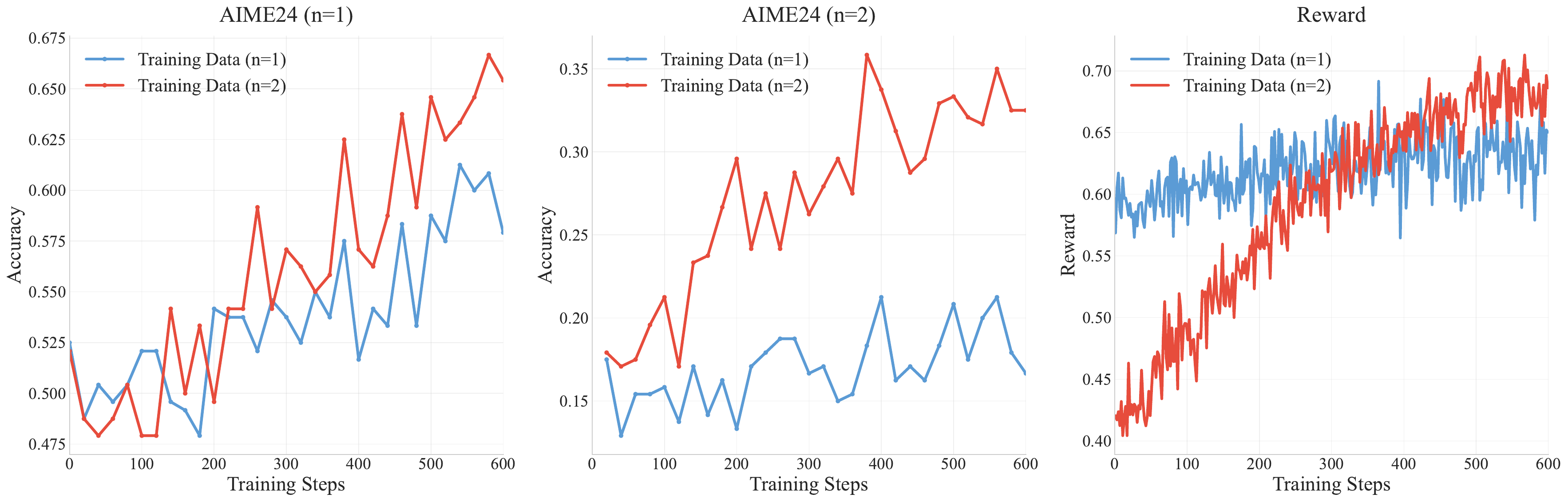}
  \caption{Training curves comparing single and composed data on $\text{AIME24}_\text{avg@8}$ and reward.}
  \label{fig:rl_n1_n2_comparison}
  \vspace{-.5em}
\end{figure*}

We train R1-Qwen-7B using both original data and 2-query composed data. As shown in Figure~\ref{fig:rl_n1_n2_comparison}, compared to the original data, composed data significantly improves performance on composed problems (\textbf{+17.4}  on AIME24 (n=2)). Additionally, we find that training with composed problem data also substantially improves performance on the original tasks (\textbf{+7.5}  on AIME24). During the training process, the reward for composed data gradually increases and surpasses the reward for the original data. 

\paragraph{Impact of Number of Composed Queries and Different Reward Schemes}
To further investigate the impact of the number of composed problems, we construct four types of training data based on the number of composed problems: composed problem counts of (1, 2, 4, and a mixture of problems with counts 1, 2, 3, 4). We also study the effects of different rewards on composed data in Table~\ref{tab:results}.

\definecolor{oursBlue}{RGB}{51,202,246}

\begin{table}[ht]
    \centering
    \caption{Results of different number of composed queries and reward function}
    \renewcommand{\arraystretch}{1.15}
    \setlength\dashlinedash{3pt}
    \setlength\dashlinegap{2pt}
    \setlength\arrayrulewidth{0.4pt}
    \resizebox{\textwidth}{!}{
    \begin{tabular}{l cc cc cc cc cc}
        \toprule
        & \multicolumn{2}{c}{\textbf{MATH500}} & \multicolumn{2}{c}{\textbf{AIME24}} & \multicolumn{2}{c}{\textbf{AIME25}} & \multicolumn{2}{c}{\textbf{AMC23}} & \multicolumn{2}{c}{\textbf{Avg.}} \\
        \cmidrule(lr){2-3} \cmidrule(lr){4-5} \cmidrule(lr){6-7} \cmidrule(lr){8-9} \cmidrule(lr){10-11}
        \textbf{Model} & \textbf{Origin} & \textbf{n=8} & \textbf{Origin} & \textbf{n=2} & \textbf{Origin} & \textbf{n=2} & \textbf{Origin} & \textbf{n=2} & \textbf{Origin} & \textbf{Multi} \\
        \midrule
        R1-Qwen-7B  & 93.6 & \cellcolor{oursBlue!12}11.8 & 48.3 & \cellcolor{oursBlue!12}16.4 & 33.3 & \cellcolor{oursBlue!12}3.5 & 90.2 & \cellcolor{oursBlue!12}48.8 & 66.4 & \cellcolor{oursBlue!12}20.1 \\
        \midrule
        Naive Training Data (n=1) & 95.6 & \cellcolor{oursBlue!12}8.4 & 57.9 & \cellcolor{oursBlue!12}16.7 & 47.9 & \cellcolor{oursBlue!12}5.1 & \textbf{95.9} & \cellcolor{oursBlue!12}55.0 & 74.3 & \cellcolor{oursBlue!12}21.3 \\
        \addlinespace[0.3ex]
        \hdashline
        \addlinespace[0.4ex]
         w/ composed queries (n=2) & 95.4 & \cellcolor{oursBlue!12}21.4 & \textbf{65.4} & \cellcolor{oursBlue!12}34.1 & \textbf{49.6} & \cellcolor{oursBlue!12}10.0 & 94.1 & \cellcolor{oursBlue!12}80.6 & \textbf{76.1} & \cellcolor{oursBlue!12}36.5 \\
         w/ composed queries (n=4) & 94.6 & \cellcolor{oursBlue!12}\textbf{50.6} & 62.9 & \cellcolor{oursBlue!12}34.8 & 45.4 & \cellcolor{oursBlue!12}8.1 & 91.9 & \cellcolor{oursBlue!12}79.1 & 73.7 & \cellcolor{oursBlue!12}\textbf{43.2} \\
         w/ composed queries (mixed) & \textbf{96.8} & \cellcolor{oursBlue!12}47.8 & 57.1 & \cellcolor{oursBlue!12}32.8 & 44.2 & \cellcolor{oursBlue!12}10.0 & 93.1 & \cellcolor{oursBlue!12}81.6 & 72.8 & \cellcolor{oursBlue!12}43.1 \\
        \addlinespace[0.3ex]
        \hdashline
        \addlinespace[0.4ex]
        w/ $R_{\text{all}}$ (n=2)  & 95.0 & \cellcolor{oursBlue!12}26.8 & 64.6 & \cellcolor{oursBlue!12}\textbf{38.8} & 48.8 & \cellcolor{oursBlue!12}\textbf{11.9} & 95.0 & \cellcolor{oursBlue!12}\textbf{83.4} & 75.9 & \cellcolor{oursBlue!12}40.2 \\
        \bottomrule
    \end{tabular}
    }
    \label{tab:results}
\end{table}

All models trained with composed data demonstrate significant performance improvements on composed problems. Moreover, composed data also substantially enhances performance on the original datasets. For instance, composed problems with n=2 yield the largest improvements on AIME24 and AIME25. As the number of composed problems increases, models exhibit stronger capabilities in handling problems requiring more reasoning steps. Additionally, we observe that using $R_{\text{all}}$ as the reward function on training data with 2 composed problems outperforms $R_{\text{last}}$ when confronting scenarios with multiple problems.
More training dynamics are provided in Appendix~\ref{appendix:training_dynamics}.

\section{Analysis}
\vspace{-.5em}
Our analysis covers evaluation results of the \name~benchmark (Section~\ref{sec:analysis:evaluation}) and RL training results using \name~datasets (Section~\ref{sec:analysis:training}). Ablation studies on evaluation metrics, dependency relationships, and problem difficulty ordering are in Appendix~\ref{appendix:ablation}.

\subsection{Evaluation Result Analysis}
\vspace{-.5em}
\label{sec:analysis:evaluation}

\begin{figure*}[ht]
  \centering
  \includegraphics[width=\textwidth]{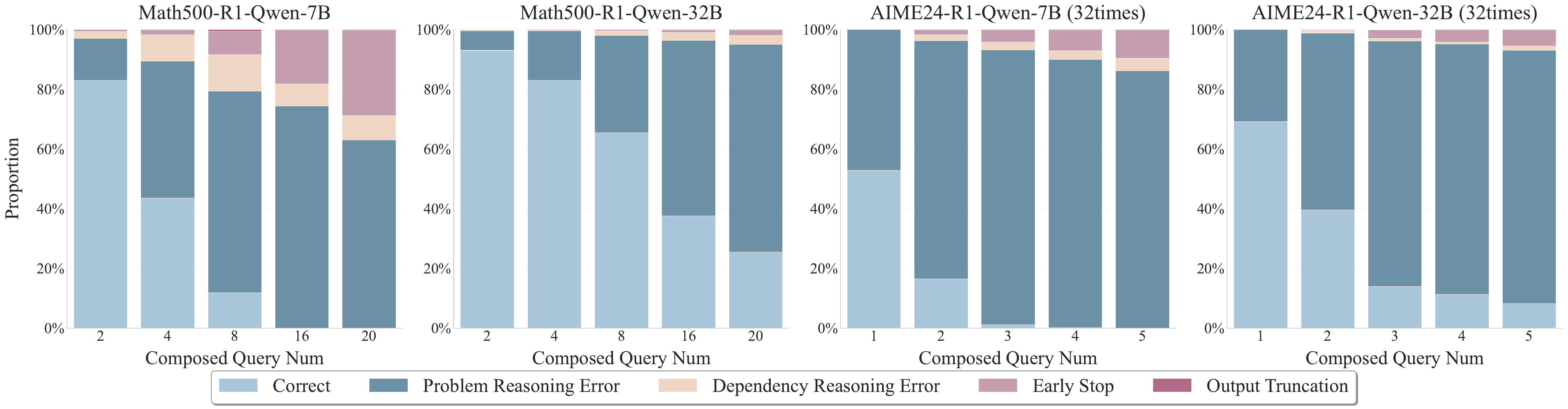}
  \caption{Error type distribution across different query numbers. 
  Four error categories: \textbf{Problem Reasoning Error} represents reasoning errors made by the model for specific problems; \textbf{Dependency Reasoning Error} indicates the model correctly solved previous problems but made errors when calculating the dependencies; \textbf{Early Stop} indicates the model prematurely terminated generation after solving previous problems; \textbf{Output Truncation} indicates generation exceeded token limit.
  }
  \label{fig:error_type_distribution}
  \vspace{-.5em}
\end{figure*}

\paragraph{Error Type Analysis}
We analyze the error types of the evaluation result in Figure~\ref{fig:error_type_distribution}. 
We find that as the number of problems increases, Problem Reasoning Errors increase rapidly. 
Adding simple dependencies between problems increases the overall reasoning difficulty, and the number of Dependency Reasoning Errors gradually increases with the number of problems, though the overall count remains relatively small. We observe that when facing multiple problems, models frequently terminate their responses prematurely, answering only a subset of the problems.

\begin{figure*}[ht]
  \centering
  \includegraphics[width=\textwidth]{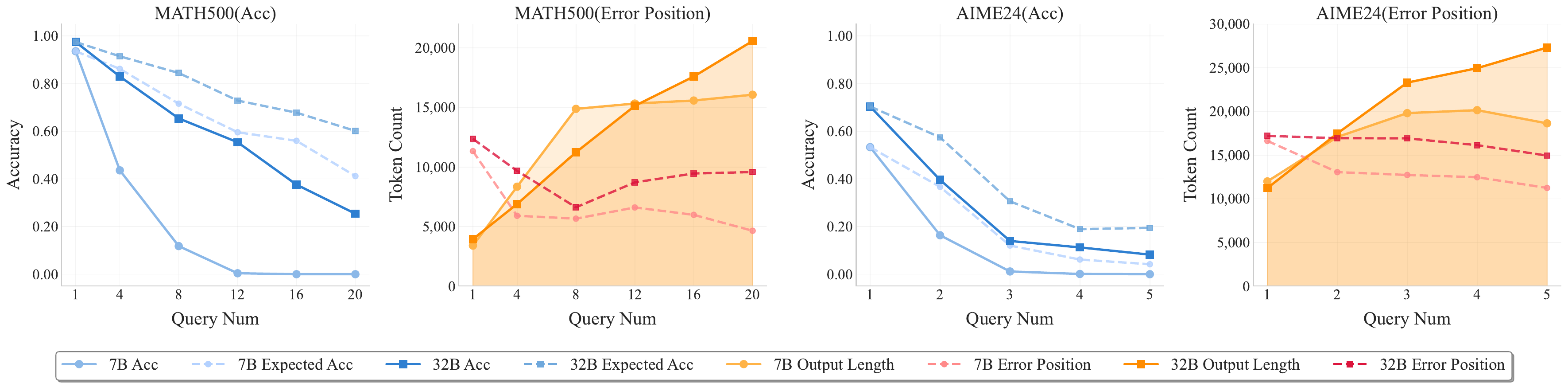}
  \caption{Analysis of accuracy and error position with R1-Qwen-7B and R1-Qwen-32B. 
  }
  \label{fig:math500_aime_analysis}
  \vspace{-1em}
\end{figure*}

\paragraph{Effective Reasoning Length of LRMs}
As shown in Figure~\ref{fig:math500_aime_analysis}, as the number of problems increases, the gap between the actual accuracy and theoretical accuracy of models becomes increasingly larger, indicating that models struggle to maintain their original performance as reasoning length increases. 
We observe that the error position of models gradually declines and stabilizes within a certain range as the number of problems increases. Comparing R1-Qwen-7B and R1-Qwen-32B, we observe that larger models can reason over longer contexts, and each model has its own reasoning boundary. 
For example, the 7B model's error range is (4-6k tokens) while the 32B model's error range is (8-10k tokens).

\begin{figure*}[ht]
  \centering
  \includegraphics[width=\textwidth]{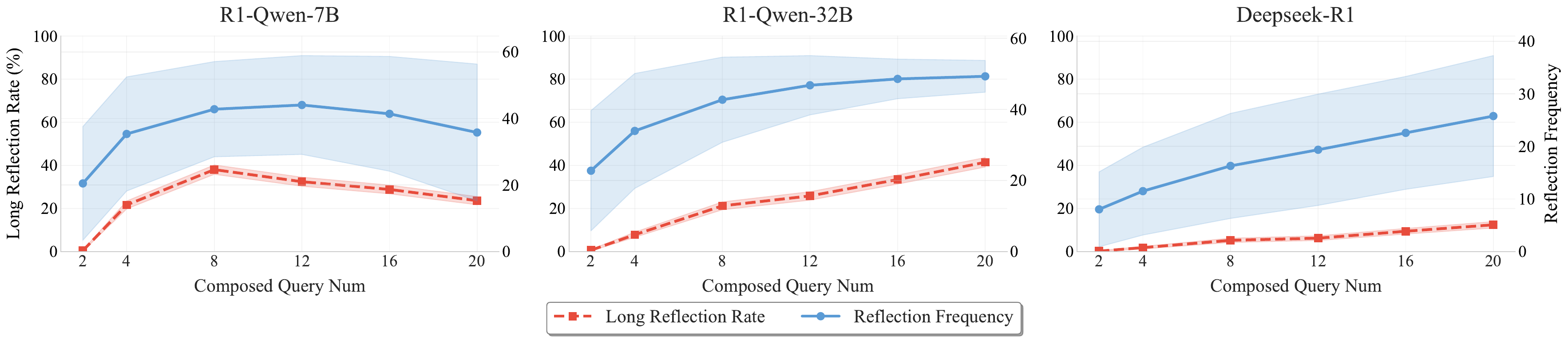}
  \caption{Reflection analysis on MATH500 dataset. Reflection Frequency refers to the average number of reflections per question. 
  Long Reflection Rate refers to the proportion of questions whose reflection range exceeds the current question.
  }
  \label{fig:math500_reflection_analysis}
  \vspace{-.5em}
\end{figure*}

\paragraph{Reflection Frequency and Depth of LRMs}
As shown in Figure~\ref{fig:math500_reflection_analysis}, the reflection frequency of models gradually increases with the number of problems and converges to a maximum value. 
As the number of problems increases, the proportion of problems involving long-range reflection also rises, yet we find that more than half of the problems lack any long-range reflection process, which indicates that LRMs' reflections are highly localized.

\begin{figure*}[ht]
  \centering
  \includegraphics[width=\textwidth]{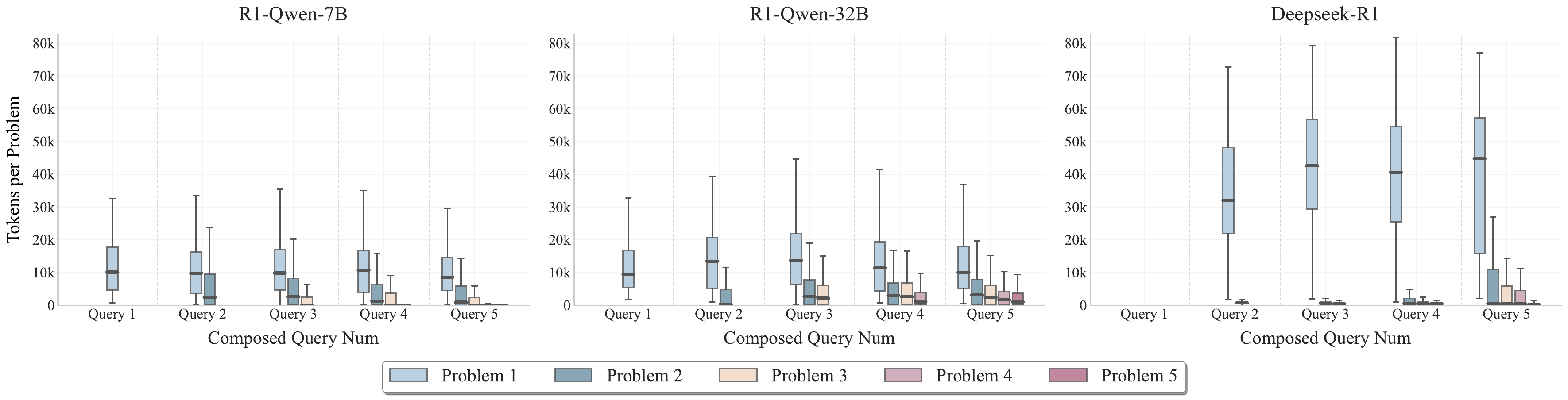}
  \caption{The thinking budget allocation for different query configurations (1-5 queries) across R1-Qwen-7B, R1-Qwen-32B, and Deepseek-R1 models on AIME24 datasets.}
  \label{fig:aime24_token_analysis}
  \vspace{-.5em}
\end{figure*}

\paragraph{Thinking Budget Allocation of LRMs}
As shown in Figure~\ref{fig:aime24_token_analysis}, current models tend to allocate more tokens to early reasoning stages. 
Even DeepSeek-R1 cannot effectively distribute the thinking budget reasonably to subsequent problems, indicating that current mainstream LRMs have not yet developed the capability to allocate thinking budgets according to reasoning horizon.

\subsection{Analysis of Reinforcement Learning with \name}
\label{sec:analysis:training}
\begin{figure*}[ht]
  \centering
  \includegraphics[width=\textwidth]{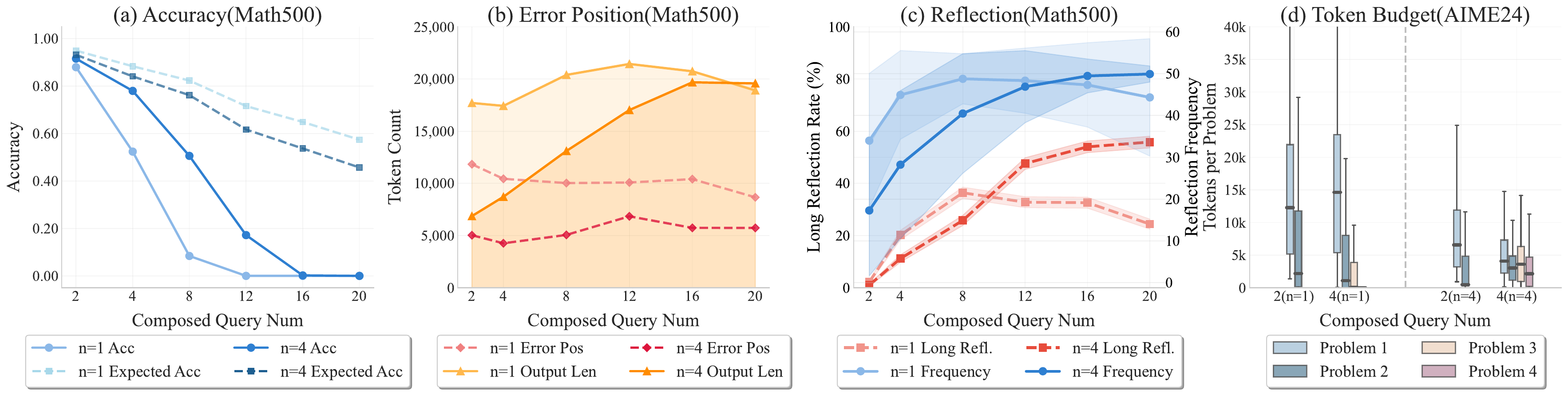}
  \caption{Analysis of reinforcement learning effects with single and composed datasets.
  (a) Math500 performance comparison, (b) error position analysis, (c) reflection analysis, and (d) token budget allocation across multi-horizon scenarios.
  }
  \label{fig:reinforcement_learning_analysis}
  \vspace{-.5em}
\end{figure*}

We analyze models trained with \name{} data versus those trained with original data using RL, as shown in Figure~\ref{fig:reinforcement_learning_analysis}. In Figure~\ref{fig:reinforcement_learning_analysis} (a), we find that training with composed queries significantly improves model performance on composed tasks and can generalize to longer reasoning horizons. Additionally, we find it alleviates the overthinking phenomenon. Models generate shorter responses when facing multiple problems compared to models trained on original data in Figure~\ref{fig:reinforcement_learning_analysis} (b), and training with composed problems enables models to learn more reasonable token budget allocation in Figure~\ref{fig:reinforcement_learning_analysis} (d). 
These results demonstrate that \textit{training with composed data promotes efficient reasoning}, which is consistent with the findings of training dynamics in Appendix~\ref{appendix:training_dynamics}.
We also provide a case study in Appendix~\ref{appendix:case_study} to compare the reasoning behavior between standard training and training with \name datasets.

In Figure~\ref{fig:reinforcement_learning_analysis} (c), as the number of problems increases, training with composed problems enables models to engage in longer reflections with increasing frequency, while the reflection frequency of models also increases more reasonably. This demonstrates that using \name~facilitates longer-range reflection in models, thereby improving performance on long-horizon reasoning tasks.

\section{Conclusion}
In this paper, we present \name, a novel and efficient approach to stimulating long-horizon reasoning in LRMs through query composition. By composing simple problems into sequential, interdependent tasks, \name{} constructs multi-step reasoning datasets that serve dual purposes: evaluating LRMs' long-horizon reasoning capabilities and enhancing their complex reasoning abilities during training. Our method establishes a foundation for future advances in complex reasoning data synthesis and the development of models with robust long-horizon reasoning capabilities.

\bibliographystyle{iclr2026_conference}

\appendix
\clearpage

\section{\name Datasets Construction for code and agentic tasks}
\paragraph{Datasets Construction for Code Tasks}
\label{appendix:dataset_construction}
For code tasks, we adopt a composition approach similar to mathematical tasks, using data points from existing datasets as seed questions for composition. We continue to employ the Expanded Problem Composition process described in Section~\ref{method:query_composition}. 
However, unlike the sequential composition used for mathematical tasks, we apply a \textit{directly composed} concatenation format for code tasks without adding explicit dependencies between problems.
This design choice is motivated by the fact that code tasks require sandbox execution to obtain answers, making it challenging to construct direct dependency relationships between problems and answers as in mathematical tasks.

\paragraph{Datasets Construction for Agentic Tasks}
For agentic tasks, we incorporate web search tasks for evaluation. We decompose questions based on the structured data from WebShaper~\citep{tao2025webshaper}, which consists of ``Target'' (target variable) and ``Variable'' (intermediate variables). Specifically, for each question, we follow this processing pipeline:
We filter the original WebShaper dataset to obtain questions with varying complexity levels, ultimately selecting 50 questions. Each question's associated URLs are accessed using a browsing tool, with browsing results stored for subsequent processing (URLs that cannot be accessed are filtered out). We employ Claude-Sonnet-4 to extract values for each variable $V$ from the web pages (variables that cannot be extracted are excluded). The original questions and variables $V$ are then assembled into a directed acyclic graph (DAG). Following topological sorting, we perform pruning to derive sub-questions and seed questions (questions with erroneous or duplicate decompositions are filtered out). 
This process yields a final dataset of 50 questions, with each question categorized into 5 levels based on the number of variables (ranging from 1 to 5), resulting in a total of 250 seed problems.

\section{How Reinforcement Learning improves long-horizon Reasoning}
\label{appendix:skywork_baseline}
Despite reinforcement learning bringing long CoT thinking capabilities to models, we find that current mainstream LRMs still cannot achieve good performance on \name{} evaluation. To further analyze the relationship between long-horizon reasoning capabilities and RL, we follow Skywork-OR1~\citep{skywork-or1-2025}, an effective and scalable RL implementation for long CoT models, to observe the changes in long-horizon reasoning capabilities of long CoT models before and after RL. 

\paragraph{Training Setup} 
We follow the Skywork OR1~\citep{skywork-or1-2025} multi-stage training approach, gradually increasing context length across different stages. Once the model's performance converged, we increased the context length in the subsequent stage. This approach led to significant performance improvements on benchmarks while also enhancing training efficiency. We employ 3 stage training with max response lengths increasing from 8k (0-600 steps) to 16k (600-1400 steps), and finally to 32k (1400-1680 steps). We train on the math subsets of the Skywork-RL dataset. Additional training settings are provided in Appendix~\ref{appendix:training_setup}.

\paragraph{Observations During Training Process}
\begin{figure*}[ht]
  \centering
  \includegraphics[width=\textwidth]{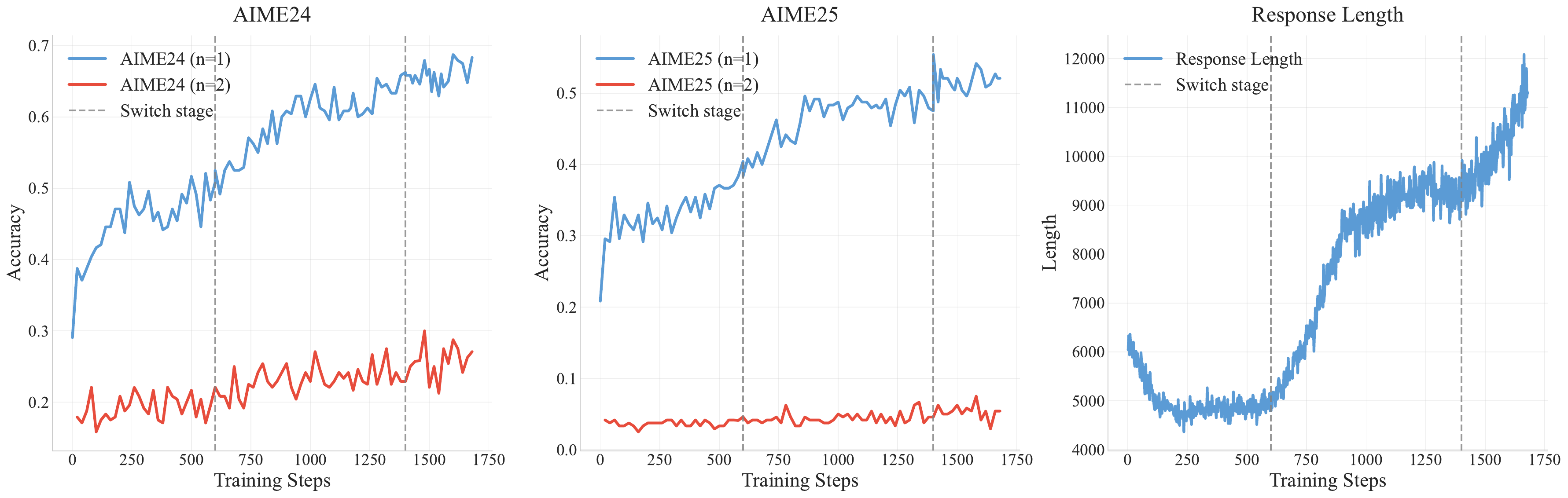}
  \caption{The AIME24, AIME25 performance for single query and 2-query settings and response length evolution during multi-stage training progression 
  across 8k, 16k, and 32k context lengths. Vertical dashed lines mark stage transitions.}
  \label{fig:multistage_training}
  \vspace{-.5em}
\end{figure*}

We find that RL training can improve model performance on composed problems, but the improvement is smaller than that on corresponding single problems (+36.6\% on AIME24 and +9.1\% on AIME24 $n=2$). Additionally, we observe that the improvement on composed problems shows no clear correlation with the increase in response length. When training at the 32k stage, although response length increases significantly, the model's performance on both single and composed problems does not improve substantially.

\section{Training Dynamics of RL with \name} 
\label{appendix:training_dynamics}
\begin{figure*}[ht]
  \centering
  \includegraphics[width=\textwidth]{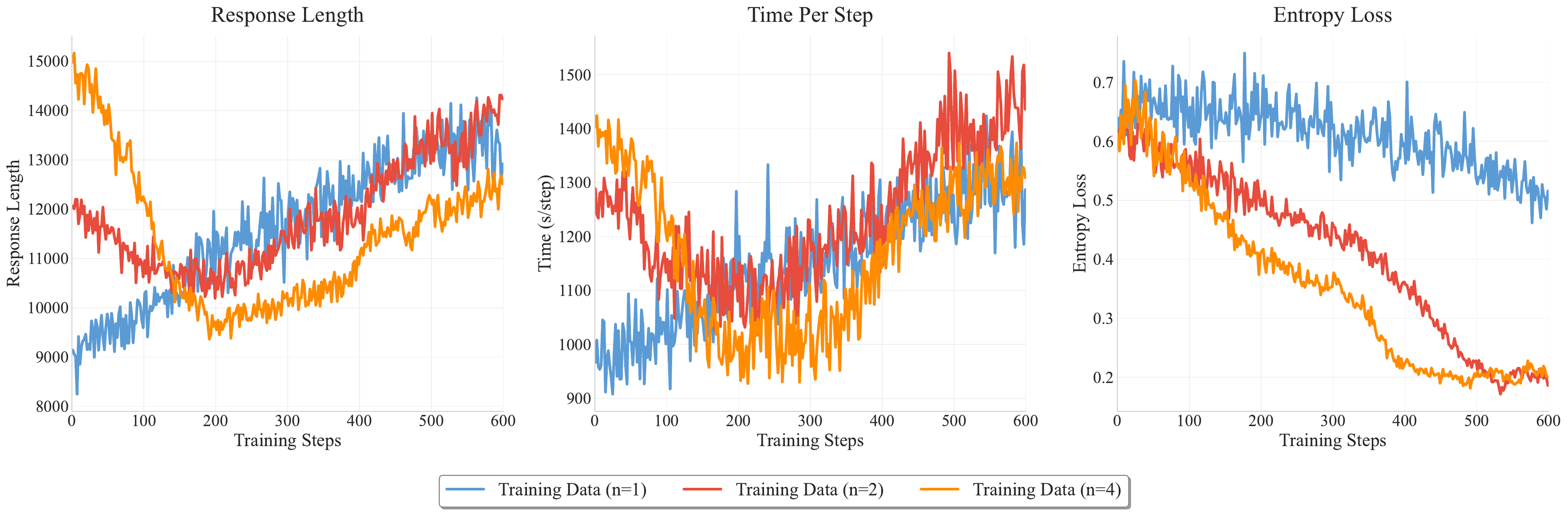}
  \caption{Training dynamics comparison across different training data compositions (n=1, n=2, n=4) showing response length, training time per step, and entropy loss evolution during the RL training process.}
  \label{fig:skywork_training_metrics}
  \vspace{-.5em}
\end{figure*}

We present the training dynamics of models trained with composed training data ($n=1, n=2, n=4$) in Figure~\ref{fig:skywork_training_metrics}. The response length of models trained with composed data initially decreases and then increases as training progresses, ultimately reaching levels comparable to those trained with original data, with similar training time per step. This indicates that models require fewer tokens to solve each problem, demonstrating that \textit{training with composed data promotes efficient reasoning}. However, the entropy loss of models trained with composed data decreases more rapidly than those trained with original data, which may limit the model's capacity for effective exploration.

\section{Ablation study}
\label{appendix:ablation}

\subsection{Ablation on Dependencies}
\label{appendix:ablation:dependencies}
\vspace{-0.5em}
\begin{wrapfigure}{r}{0.35\textwidth}
  \centering
  \vspace{-6mm}
  \includegraphics[width=0.35\textwidth]{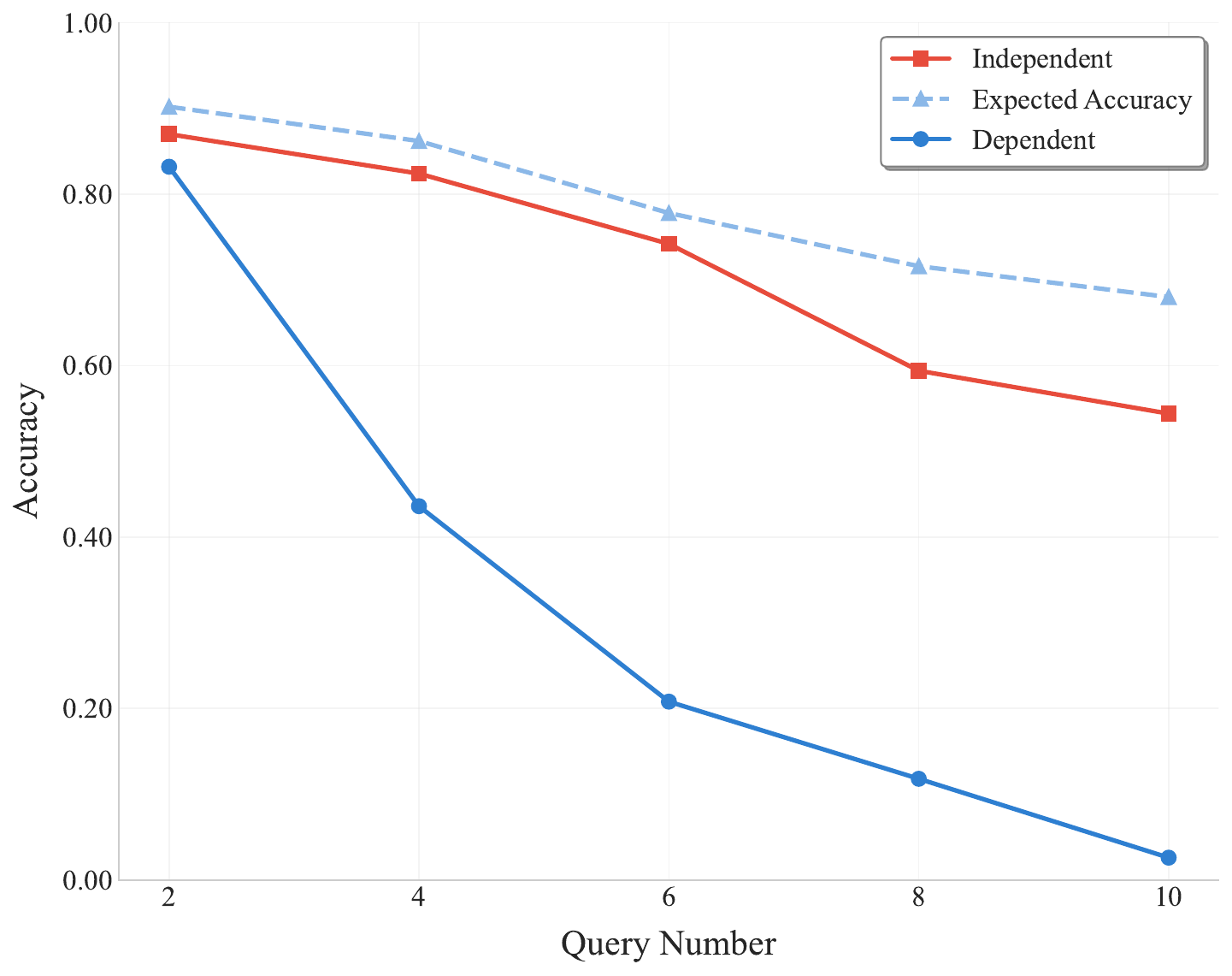}
  \vspace{-6mm}
  \caption{Comparison between multiple dependent and independent problems.}
  \label{fig:dependent_vs_independent}
  \vspace{-.8em}
\end{wrapfigure}
We compare the difference between multiple dependent problems and multiple independent problems. We remove the dependency construction step and directly concatenate multiple problems. We conduct experiments using R1-Qwen-7B on Math500, with results shown in Figure~\ref{fig:dependent_vs_independent}. We find that the accuracy of both problem composition methods falls below the theoretical accuracy, and the accuracy of multiple sequentially dependent problems is significantly lower than that of multiple independent problems. This indicates that current models still have substantial deficiencies when handling multiple correlated problems.

\subsection{Ablation on Evaluation Metric}
\label{appendix:ablation:metric}
\begin{figure*}[ht]
  \centering
  \includegraphics[width=\textwidth]{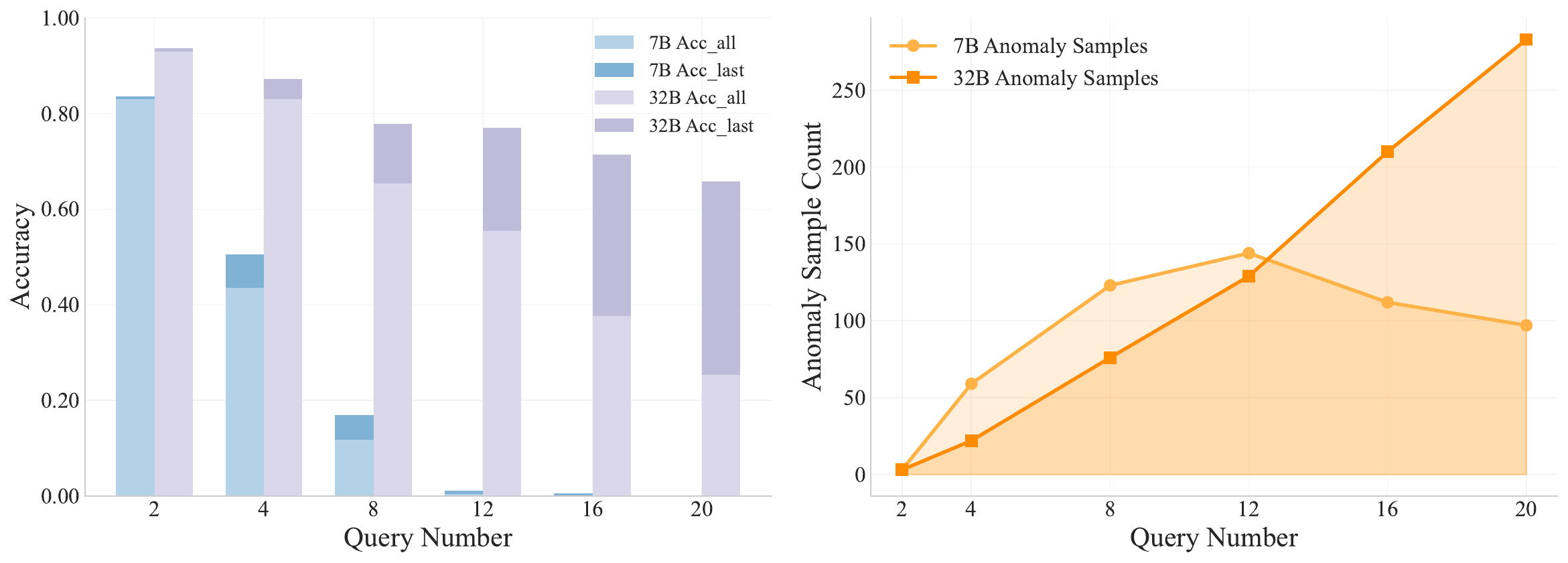}
  \caption{R1-Qwen models showing anomalous behavior in sequential reasoning. Left: $\text{Acc}_{\text{all}}$ vs. $\text{Acc}_{\text{last}}$ revealing increasing divergence. Right: Anomalous sample counts where models correctly answer final problems despite preceding errors.}
  \label{fig:anomaly_analysis}
  \vspace{-.5em}
\end{figure*}

\name{} adopts an all-or-nothing scoring criterion $\text{Acc}_{\text{all}}$ to ensure models correctly answer all problems. An alternative evaluation metric $\text{Acc}_{\text{last}}$ considers a response correct if only the final problem is answered correctly. Theoretically, these two metrics should be identical for problems with sequential dependencies, as correctly answering the final problem requires sequentially solving all preceding problems. However, our ablation experiments reveal substantial differences between these metrics as the number of problems increases, as shown in Figure~\ref{fig:anomaly_analysis} (Left). The probability of correctly answering only the final problem far exceeds the probability of correctly answering all problems. We observe an anomalous phenomenon: models can correctly solve subsequent problems despite incorrect solutions to preceding ones, indicating that models can produce correct answers even when problems should be unsolvable. We provide statistics on these anomalous cases in Figure~\ref{fig:anomaly_analysis} (right). We hypothesize that this phenomenon is related to data contamination in models~\citep{wu2025reasoningmemorizationunreliableresults}.




\subsection{Impact of Query Difficulty Ordering}
\label{appendix:ablation:difficulty}

\begin{figure*}[ht]
  \centering
  \includegraphics[width=\textwidth]{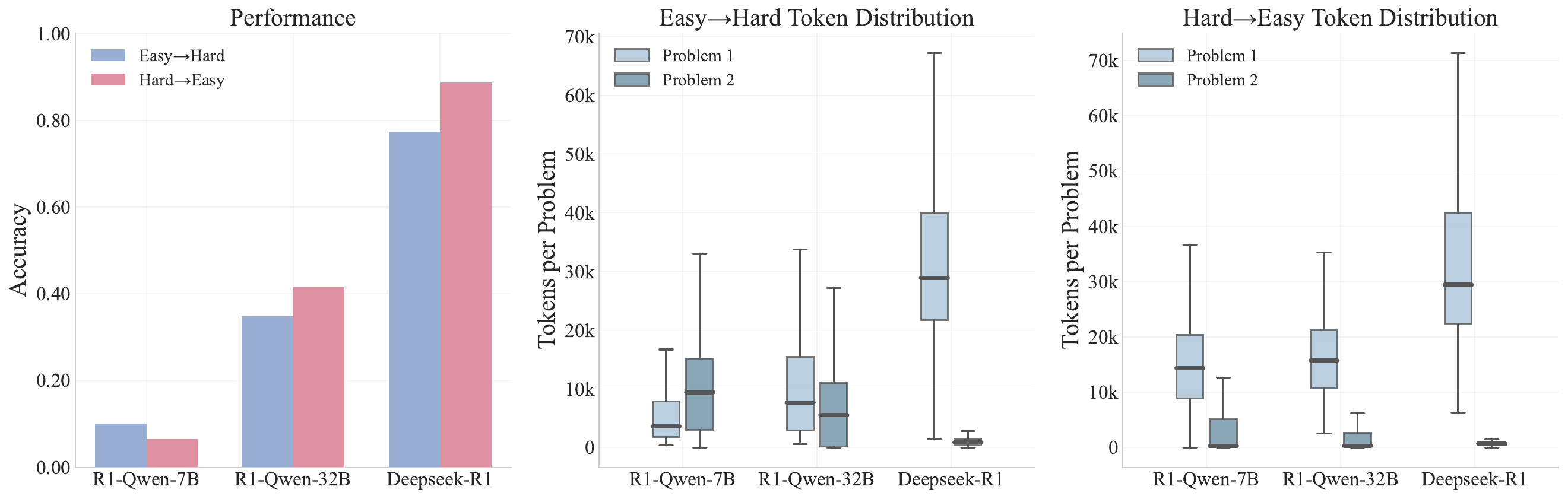}
  \caption{Ablation study on the impact of query difficulty ordering for R1-Qwen-7B, R1-Qwen-32B, and DeepsSeek-R1 models. (a) Performance comparison between easy-to-hard and hard-to-easy query orderings. (b) Thinking budget allocation in the easy-to-hard scenario. (c) Thinking budget allocation in the hard-to-easy scenario.}
  \label{fig:aime24_query_ordering_analysis}
  \vspace{-.5em}
\end{figure*}

We conduct an ablation study to examine whether the ordering of query difficulty affects model performance and thinking budget allocation. 
Using the pass rate of R1-Qwen-7B as the reference metric, we define a query as \textit{easy} if its pass rate exceeds 0.5 and \textit{hard} otherwise. 
We then compare the performance of both 7B and 32B models under different orderings of easy and hard queries (i.e., easy-to-hard vs.~hard-to-easy). 
Figure~\ref{fig:aime24_query_ordering_analysis} (b) and (c) show that all models fail to allocate thinking budget reasonably according to problem difficulty. More powerful models (DeepSeek-R1, R1-Qwen-32B) can benefit from difficulty ordering in Figure~\ref{fig:aime24_query_ordering_analysis} (a). We hypothesize that this is because placing difficult problems at the beginning leads models to allocate more token budget to difficult problems, thereby improving overall success rate, while smaller models (R1-Qwen-7B) show no significant benefit.

\section{Evaluation Implementation Details}
\subsection{Models and Datasets in \name Benchmark}
\label{appendix:model_datasets_details}
\paragraph{Datasets Statistics and Evaluation Metric}
We present the statistics and evaluation metric of the \name~benchmark in Table~\ref{tab:dataset_stats}, showing the number of problems in the original datasets, extracted seed questions, and final composed datasets.

\begin{table}[ht]
\centering
\caption{Dataset statistics and evaluation metric for \name~benchmark}
\label{tab:dataset_stats}
\begin{tabular}{@{}lcccc@{}}
\toprule
\multirow{2}{*}{\textbf{Dataset}} & \multicolumn{3}{c}{\textbf{Number of Problems}} & \multirow{2}{*}{\textbf{Metric}} \\
\cmidrule(lr){2-4}
& \textbf{Original} & \textbf{Seed} & \textbf{Composed} & \\
\midrule
\rowcolor{gray!10} \multicolumn{5}{c}{\textit{Mathematical Tasks}} \\
Math500 & 500 & 257 & 500 & Accuracy \\
AIME24 & 30 & 28 & 30 & Avg@32 \\
AIME25 & 30 & 28 & 30 & Avg@32 \\
AMC23 & 40 & 37 & 40 & Avg@8 \\
\midrule
\rowcolor{gray!10} \multicolumn{5}{c}{\textit{Code Tasks}} \\
LiveCodeBench & 279 & 279 & 279 & Pass@1 \\
\midrule
\rowcolor{gray!10} \multicolumn{5}{c}{\textit{Agentic Tasks}} \\
WebShaper & 500 & 117 & 50 & Avg@3 \\
\bottomrule
\end{tabular}
\end{table}

\paragraph{Model Details} 
In the \name~benchmark, we evaluate the following open-source models. We present the model sources and their corresponding evaluation lengths (max new tokens for generation) as follows: DeepSeek-R1-0528 (64k), R1-Qwen-1.5B (64k), R1-Qwen-7B (64k), R1-Qwen-32B (64k), R1-Llama8B (64k), R1-Llama70B (64k)~\citep{guo2025deepseek}, Qwen3-8B (40k), Qwen3-32B (40k), Qwen3-235B-A22B-2507 (64k), Qwen3-235B-A22B-Thinking-2507 (64k), QwQ-32B (64k)~\citep{yang2025qwen3technicalreport}, Nemotron-Research-Reasoning-Qwen-1.5B (64k), Llama-3.1-Nemotron-Nano-8B-v1 (64k)~\citep{nvidia2024nemotron4340btechnicalreport}, DeepScaleR-1.5B-Preview (64k)~\citep{deepscaler2025}, Polaris-1.7B-Preview (64k), Polaris-4B-Preview (64k)~\citep{Polaris2025}, Skywork-OR1-7B (64k), Skywork-OR1-32B (64k)~\citep{skywork-or1-2025}, OpenThinker3-7B (32k)~\citep{guha2025openthoughtsdatarecipesreasoning}, Efficient-R1-7B ($\alpha = 0.2$) (64k)~\citep{arora2025traininglanguagemodelsreason}, Laser-DE-L4096-7B (64k)~\citep{liu2025learnreasonefficientlyadaptive}, DAPO-Qwen-32B (64k)~\citep{yu2025dapoopensourcellmreinforcement}.

\paragraph{Prompt Examples} 
We present the prompt examples for math, code, and web search tasks in Figure~\ref{fig:math_prompt}, Figure~\ref{fig:code_prompt} and Figure~\ref{fig:agent_prompt}.

\begin{figure}[ht]
\begin{prompt}{\text{Prompt Example for Mathematical Tasks}}{math_prompt}
\small
Problem 1: \{Question 1\}
Using the result \{[answer 1]\} from the previous calculation, \{[variable 2]\} = \{dependency\_function\}. 
Problem 2: \{Question 2\}
......
Problem n: \{Question n\}
Note: In this problem set:
- [variable k] represents the calculated variable needed to solve problem k.
- [answer k] represents the answer to problem k.
Solve all problems step by step and provide the answers for all problems in the following format:
\#\#\# Final Answers
Problem 1: \textbackslash boxed\{[answer 1]\}
Problem 2: \textbackslash boxed\{[answer 2]\}
......
Problem n: \textbackslash boxed\{[answer n]\}

\end{prompt}
\caption{Prompt Example for Mathematical Tasks}
\label{fig:math_prompt}
\end{figure}

\begin{figure}[ht]
\begin{prompt}{\text{Prompt Example for Code Tasks}}{code_prompt}
\small
\#\#\# Question:
\{Questions\}
Solve all problems step by step and provide the code for all problems. For each problem, read the inputs from stdin solve the problem and write the answer to stdout (do not directly test on the sample inputs). Enclose your code within delimiters in the following format:
```python \# YOUR CODE HERE ```
\#\#\# Answer: (use the provided format with backticks)

\end{prompt}
\caption{Prompt Example for Code Tasks}
\label{fig:code_prompt}
\end{figure}

\begin{figure}[ht]
\begin{prompt}{\text{Prompt Example for Web Search Tasks}}{agent_prompt}
\small
You may call one or more functions to assist with the user query.\textbackslash n\textbackslash nYou are provided with function signatures within <tools></tools> XML tags:
\{Tools\}
Please answer the following question by using the available tools to search for information and visit relevant webpages. Before each action, conduct brief reasoning to determine your next step. After reasoning, if you find you lack some knowledge, you can call the tools to get new information.
If you find no further external knowledge needed, you can provide the answer within <answer></answer> XML tags. For example, <answer> Beijing </answer>.
\{Question\}
\end{prompt}
\caption{Prompt Example for Web Search Tasks}
\label{fig:agent_prompt}
\end{figure}

\subsection{Evaluation Metrics Calculation}
\label{appendix:evaluation_metrics}
For mathematical and agent-based WebShaper tasks, we utilize GPT-4.1 to extract answers from all problems and perform subsequent scoring. For code tasks, we first extract code blocks from the responses and assess their correctness via sandbox execution. The prompts used for scoring are presented in Figure~\ref{fig:math_prompt_extraction} and Figure~\ref{fig:agent_prompt_extraction}.

\begin{figure}[ht]
\begin{prompt}{\text{Answer Extraction Prompt for Mathematical Tasks}}{math_prompt_extraction}
\small
You are a helpful assistant. Extract the question numbers and corresponding answers from the given results, and return them in JSON format, where the key is the question number and the value is the answer.

\end{prompt}
\caption{Answer Extraction Prompt for Mathematical Tasks}
\label{fig:math_prompt_extraction}
\end{figure}

\begin{figure}[ht]
\begin{prompt}{\text{Answer Extraction Prompt for WebShaper}}{agent_prompt_extraction}
\small
You are an evaluation assistant. Please determine if the predicted answer is equivalent to the labeled answer.

Question: \{question\}

Labeled Answer: \{correct\_answer\}

Predicted Answer: \{response\}

Did the model give an answer equivalent to the labeled answer? Please respond with ``Correct'' if they are equivalent, or ``Incorrect'' if they are not equivalent. Do not include any other text.
\end{prompt}
\caption{Answer Extraction Prompt for WebShaper}
\label{fig:agent_prompt_extraction}
\end{figure}

We also compare the consistency rate between using model-based answer extraction and rule-based ``\textbackslash boxed\{\}'' pattern extraction in Table~\ref{tab:extraction}. We find that as the number of problems increases, many models fail to accurately follow the output format, making model-based answer extraction more accurate for evaluation. Therefore, we uniformly adopt model-based answer extraction for mathematical tasks.
\begin{table}[ht]
\centering
\caption{Consistency rate between model-based and rule-based extraction for R1-Qwen-7B on Math500}
\label{tab:extraction}
\begin{tabular}{@{}lcccc@{}}
\toprule
\textbf{Composed Problem Num} & \textbf{2} & \textbf{4} & \textbf{8} & \textbf{16} \\
\midrule
Consistency Rate (\%) & 96.83 & 96.41 & 93.77 & 91.04 \\
\bottomrule
\end{tabular}
\end{table}

\subsection{Inference Hyperparameters}
\label{appendix:evaluation_inference_setting}
We set the maximum generation length for inference to 64k tokens. For models with maximum lengths below 64k, we set the max generation length to their max sequence length. 
For inference hyperparameters, we set temperature to 1.0, top-$k$ to 10, and top-$p$ to 0.95. For the Qwen series hybrid reasoning models that switch between thinking mode and non-thinking mode, we consistently test their thinking mode.

\section{Training Implementation Details}
\label{appendix:training_setup}
\subsection{Training Setup}
\label{appendix:training_skywork}
We show the training hyperparameters for training with \name datasets in Section~\ref{sec:training_horizon}. The maximum response length is set to 40k tokens to prevent truncation.
Training is conducted exclusively on the mathematical components of the Skywork-RL dataset.
All training progress are fine-tuned by optimizing the policy loss~\ref{eq:GRPO-obj} with a constant learning rate of $1 \times 10^{-6}$. We set the batch size to 256, mini-batch size to 128, and group size to 16. We employ a higher clip ratio of 0.265, target entropy of 0.2, sampling temperature of 1.0, and rejection sampling. Notably, we do not apply any KL loss in our training process.

We use the same training hyperparameters for standard RL training in Appendix~\ref{appendix:skywork_baseline}. 
We implement a three-stage training paradigm following Skywork OR1~\citep{skywork-or1-2025}, where context length is incrementally expanded upon reaching performance convergence at each stage. This progressive approach, advancing from 8k to 16k and ultimately to 32k maximum response tokens, delivers both improved benchmark results and enhanced computational efficiency.

\subsection{\name Training Datasets}
\label{appendix:training_horizon}
We initialize a filtered data pool $\mathcal{D}_{\text{filtered}}$ from the original Skywork-OR1-RL training data via the \name{} Problem Filtering process (Section~\ref{method:problem_filtering}). To control the problem difficulty, we compose problems according to their pass rates while maintaining $\text{Acc}_{\text{expected}} > 0.25$ for all composed instances. We show the datasets' statistics in Table~\ref{tab:training_datasets_statistics}.
\begin{table}[ht]
\centering
\caption{Dataset statistics for Skywork-o1-RL Data}
\begin{tabular}{lccc}
\toprule
 & Original & Seed & Composed (pass\_rate$>$0.25) \\
\midrule
Skywork-o1-RL Data & 48371 & 18015 & 18000 \\
\bottomrule
\end{tabular}
\label{tab:training_datasets_statistics}
\end{table}

\section{The use of large language models}
Large language models were employed exclusively as writing aids to refine sentence clarity, format tables, and improve overall readability. They were not involved in the central research contributions, experimental design, or scientific content of this work. The authors bear full responsibility for all content presented in the paper.

\section{Case Study}
\label{appendix:case_study}
We provide a case study with an example prompt shown in Figure~\ref{fig:case_study_prompt}, and compare model outputs on multi-horizon problems when trained with original data versus \name~training data, as illustrated in Figure~\ref{fig:case_study_k1} and Figure~\ref{fig:case_study_k4}. We observe that models trained with \name~training data consume fewer tokens per problem, avoid excessive thinking budget allocation on individual problems, and successfully solve all problems.

\begin{figure}[ht]
\begin{prompt}{\text{Example Prompt for Case Study}}{case_study_prompt}
\small
\textcolor{blue}{\textbf{Instructions:}}

Problem 1: Alice and Bob play the following game. A stack of $n$ tokens lies before them. The players take turns with Alice going first. On each turn, the player removes either $1$ token or $4$ tokens from the stack. Whoever removes the last token wins. Find the number of positive integers $n$ less than or equal to $2024$ for which there exists a strategy for Bob that guarantees that Bob will win the game regardless of Alice's play.

Problem 2: Using the result [answer1] from the previous calculation, [variable2] = [answer1] - 805. Every morning Aya goes for a $9$-kilometer-long walk and stops at a coffee shop afterwards. When she walks at a constant speed of $s$ kilometers per hour, the walk takes her [variable2] hours, including $t$ minutes spent in the coffee shop. When she walks $s+2$ kilometers per hour, the walk takes her 2 hours and 24 minutes, including $t$ minutes spent in the coffee shop. Suppose Aya walks at $s+\frac{1}{2}$ kilometers per hour. Find the number of minutes the walk takes her, including the $t$ minutes spent in the coffee shop.

Problem 3: Using the result [answer2] from the previous calculation, [variable3] = [answer2] - 150. Let $\mathcal{B}$ be the set of rectangular boxes with surface area $[variable3]$ and volume $23$. Let $r$ be the radius of the smallest sphere that can contain each of the rectangular boxes that are elements of $\mathcal{B}$. The value of $r^2$ can be written as $\frac{p}{q}$, where $p$ and $q$ are relatively prime positive integers. Find $p+q$.

Problem 4: Using the result [answer3] from the previous calculation, [variable4] = [answer3] - 720. Let $\omega \neq [variable4]$ be a 13th root of unity. Find the remainder when
\[\prod_{k=0}^{12}(2-2\omega^k+\omega^{2k})\]
is divided by 1000.

Note: In this problem set:
- [variablek] represents the calculated variable needed to solve problem k.
- [answerk] represents the answer to problem k.

Solve all problems step by step and provide the answers for all problems in the following format:

\textbf{Final Answers}

Problem 1: \boxed{[answer1]}

Problem 2: \boxed{[answer2]}

Problem 3: \boxed{[answer3]}

Problem 4: \boxed{[answer4]}

\textcolor{blue}{\textbf{Target Answers:}}

Problem 1: \boxed{809}

Problem 2: \boxed{204}

Problem 3: \boxed{721}

Problem 4: \boxed{321}

\end{prompt}
\caption{Example Prompt for Case Study}
\label{fig:case_study_prompt}
\end{figure}

\begin{figure}[ht]
\begin{prompt}{\text{Model Output for Model Trained with Original Data}}{case_study_k1}
\small

\textcolor{blue}{\textbf{<Problem 1 Start>}}

Okay, so I have four problems here, and I need to solve each one step by step. 
Let me start with Problem 1.

\textbf{Problem 1: Game Theory}

Alice and Bob play a game with a stack of n tokens. They take turns removing 
either 1 or 4 tokens. Alice goes first, and the player who takes the last token 
wins. I need to find how many $n \leq 2024$ where Bob has a winning strategy, 
regardless of Alice's moves.

\textit{...... Problem 1 thinking}

Therefore, our answer for Problem1 is 809.

\textbf{Final Answer}

Problem 1: $\boxed{809}$

\textcolor{blue}{\textbf{<Problem 1 END, Output Length=3154, Status=Correct>}}

\textcolor{blue}{\textbf{<Problem 2 Start>}}

\textbf{Problem 2: Walking Speeds}

Every morning, Aya walks 9 km and stops at a coffee shop. Walking at speed 
$s$ km/h takes her $t$ minutes in the shop. When she walks $s+2$ km/h, the walk 
takes 2 hours and 24 minutes, including $t$ minutes. Find the number of minutes 
the walk takes when she walks at $s+1/2$ km/h.

Wait, let me try to parse this.

\textit{...... Problem 2 thinking}

When walking at $s +\frac{1}{2}$, total time=$\frac{9}{s +0.5} + \frac{t}{60}$.

But without knowing $s$ and $t$, I can't compute it.

I think I've spent too much time without progress, so I might need to move to problem3 and return.

\textcolor{blue}{\textbf{<Problem 2 END, Output Length=14055, Status=Wrong>}}

\textcolor{blue}{\textbf{<Problem 3 Start>}}

\textbf{Problem3: }

Using the result [variable2] from problem1=$809$ and [answer2] from problem2= let's say it's $100$, find the remainder when [variable3] is divided by $1000$.

\textit{...... Problem 3 thinking}

But I'm not confident.

Wait, alternatively, perhaps problem2's answer is the number of solutions $\leq 2024$, which was $809$, so problem3 uses that.

But without more info, I can't compute. Given the time constraints, I think I'll settle on:

Problem1: $809$

Problem2: $100$

Problem3: $(809 + 100) = 909$

\textcolor{blue}{\textbf{<Problem 3 END, Output Length=16055, Status=Wrong>}}

\textcolor{gray}{\textbf{<Problem 4 Start>}}

\textbf{Status:} NOT\_ATTEMPTED

\textcolor{gray}{\textbf{<Problem 4 END, Output Length=16055, Status=NOT\_ATTEMPTED>}}

\end{prompt}
\caption{Case Study for Model Trained with Original Data}
\label{fig:case_study_k1}
\end{figure}

\begin{figure}[ht]
\begin{prompt}{\text{Model Output for Model Trained with Composed Data}}{case_study_k4}
\small

\textcolor{blue}{\textbf{<Problem 1 Start>}}

Alright, I have four problems to solve. They're all math problems, each building on the previous one with some variables. Let me take them one by one.

\textbf{Problem 1:} Alice and Bob play a game with a stack of $n$ tokens. They take turns removing 1 or 4 tokens, Alice goes first. We need to find the number of positive integers $n \leq 2024$ where Bob has a winning strategy regardless of Alice's play.

This is a combinatorial game theory problem. Looking at losing positions, I find they are $n \equiv 0$ or $2 \pmod{5}$.

\textit{...... Problem 1 thinking}

So the answer for Problem 1 is 809. So answer1=809.

\textcolor{blue}{\textbf{<Problem 1 END, Output Length=2015, Status=Correct>}}

\textcolor{blue}{\textbf{<Problem 2 Start>}}

\textbf{Problem 2:} Using the result [answer1] from the previous calculation, [variable2] = [answer1] - 805. Every morning Aya goes for a 9-kilometer-long walk and stops at a coffee shop afterwards.

So variable2 = $809 - 805 = 4$. 

\textit{...... Problem 2 thinking}

Solving, I get $s = 2.5$ km/h and $t = 24$ minutes.

When she walks at $s + \frac{1}{2} = 3$ km/h: Total time = 204 minutes.

So the answer is 204. So answer2=204.

\textcolor{blue}{\textbf{<Problem 2 END, Output Length=3081, Status=Correct>}}

\textcolor{blue}{\textbf{<Problem 3 Start>}}

\textbf{Problem 3:} Using the result [answer2] from the previous calculation, [variable3] = [answer2] - 150. Let $\mathcal{B}$ be the set of rectangular boxes with surface area [variable3] and volume $23$.

So variable3 = $204 - 150 = 54$. 

\textit{...... Problem 3 thinking}

Thus $r^2 = \frac{657}{64}$, so $p = 657$, $q = 64$, $p + q = 721$.

So answer3=721.

\textcolor{blue}{\textbf{<Problem 3 END, Output Length=6819, Status=Correct>}}

\textcolor{blue}{\textbf{<Problem 4 Start>}}

\textbf{Problem 4:} Using the result [answer3] from the previous calculation, [variable4] = [answer3] - 720. Let $\omega \neq$ [variable4] be a 13th root of unity.

So variable4 = $721 - 720 = 1$. 

\textit{...... Problem 4 thinking}

The remainder when $8321$ is divided by $1000$ is $321$.

So answer4=321.

\textcolor{blue}{\textbf{<Problem 4 END, Output Length=9336, Status=Correct>}}

\end{prompt}
\caption{Case Study for Model Trained with Composed Data}
\label{fig:case_study_k4}
\end{figure}

\end{document}